\documentclass[10pt,twocolumn,letterpaper]{article}

\usepackage[accsupp]{axessibility} 
\usepackage{iccv}
\usepackage{times}
\usepackage{epsfig}
\usepackage{graphicx}
\usepackage{amsmath}
\usepackage{amssymb}

\usepackage{multirow}
\usepackage{booktabs} 

\newcommand{\minisection}[1]{\vspace{0.04in} \noindent {\bf #1}\ \ }
\iftrue

\newcommand{\alertJW}[1]{{\color{magenta}{\bf JW:#1}}}
\else
\newcommand{\alertJW}[1]{}
\fi
\usepackage[pagebackref=true,breaklinks=true,letterpaper=true,colorlinks,bookmarks=false]{hyperref}

\iccvfinalcopy 

\def\iccvPaperID{18} 

\ificcvfinal\fi

\begin{document}

\title{Reducing Label Effort: Self-Supervised meets Active Learning}
\author{Javad Zolfaghari Bengar$^{1,2}$
\and Joost van de Weijer$^{1,2}$
\and Bartlomiej Twardowski$^{1}$
\and Bogdan Raducanu$^{1,2}$\\
Computer Vision Center (CVC)$^1$, Univ. Aut\`{o}noma of Barcelona (UAB)$^2$\\

{\tt\small \{jzolfaghari,joost,btwardowski,bogdan\}@cvc.uab.es}
}
\maketitle
\ificcvfinal\thispagestyle{empty}\fi

\begin{abstract}
Active learning is a paradigm aimed at reducing the annotation effort by training the model on actively selected informative and/or representative samples. 
Another paradigm to reduce the annotation effort is self-training that learns from a large amount of unlabeled data in an unsupervised way and fine-tunes on few labeled samples. Recent developments in self-training have achieved very impressive results  rivaling supervised learning on some datasets. The current work focuses on whether the two paradigms can benefit from each other. We studied object recognition datasets including CIFAR10, CIFAR100 and Tiny ImageNet with several labeling budgets for the evaluations.  Our experiments reveal that self-training is remarkably more efficient than active learning at reducing the labeling effort, that for a low labeling budget, active learning offers no benefit to self-training, and finally that the combination of active learning and self-training is fruitful when the labeling budget is high. The performance gap between active learning trained either with self-training or from scratch diminishes as we approach to the point where almost half of the dataset is labeled.
\end{abstract}


\section{Introduction}
Deep learning methods obtain excellent results on large annotated datasets~\cite{krizhevsky2012imagenet}. However, labeling large amounts of data is labor-intensive and can be very costly. Therefore, the field of active learning explores algorithms that reduce the amount of labeled data that is required. This is achieved by labeling those unlabeled data samples (from the unlabeled data pool) that are considered most useful for the machine learning algorithm. The field of active learning can be roughly divided into two subfields. Informativeness-based methods aim to identify those data samples for which the algorithm is most uncertain~\cite{gu2014modelchange,yang2015multi,guo2010nips}. Adding these samples to the labeled data pool is expected to improve the algorithm. Representativeness-based methods aim to label data in such a way that for all unlabeled data there is a ‘representative’ (defined based on distance in feature space) labeled sample~\cite{dutt2016active,sener2018active}. 
Active learning methods are typically evaluated by supervised training of the network on only the labeled data pool: the active learning method that obtains the best results, after a number of training cycles with a fixed label budget, is then considered superior. 

Self-supervised learning of representation for visual data has seen stunning progress in recent years~\cite{chen2020SimCLR,chen2020MOCOv2,chen2021SimSiam,NEURIPS2020BYOL,he2020MOCO}, with some unsupervised methods being able to learn representations that rival those learned supervised. The main progress has come from a recent set of works that learn representations that are invariant with respect to a set of distortions of the input data 
 (such as cropping, applying blur, flipping, etc). In these methods, two distorted versions, called views, of the image are produced. Then the network is trained by enforcing the representations of the two views to be similar. To prevent these networks to converge to a trivial solution different approaches have been developed~\cite{NEURIPS2020BYOL,zbontar2021barlow}. The resulting representations are closing the gap with supervised-learned representation.  For some downstream applications, such as segmentation and detection, the self-supervised representations even outperform the supervised representations~\cite{zhao2020makes}.

As discussed, self-supervised learning can learn high-quality features that are almost at par with the features learned by supervised methods. As such it has greatly improved the usefulness of unlabeled data. The standard active learning paradigm trains an algorithm on the labeled data set, and based on the resulting algorithm selects data points that are expected to be most informative for the algorithm in better understanding the problem~\cite{settles2012active}. In this standard setup, the unlabeled data is not exploited to improve the algorithm. Given the huge performance gains that are reported by applying self-supervised learning, we propose to re-evaluate existing active learning algorithms in this new setting where the unlabeled data is exploited by employing self-supervised learning. 



Self-supervised learning and active learning both aim to reduce the label-effort. Based on our experiments we conclude the following:

\begin{itemize}
    \item In our evaluations on three datasets, Self-training is much more efficient than AL in reducing the labelling effort.
    
    \item Self-training + AL substantially outperforms AL methods. However, the performance gap diminishes for large labeling budget (approximately $50\%$ of the dataset in our experiments).
    
    \item Based on results of three datasets, Self-training+AL marginally outperforms self-training but only when the labeling budget is high.
\end{itemize}
In general, our results suggest that self-supervised learning techniques are more efficient than active learning to reduce the label effort. A small additional boost can be obtained from active learning when reaching the high label budget.

Our paper is organized as follows: In section~\ref{sec:RW} we describe the related work. Next, in section ~\ref{sec:prelim} we introduce the proposed framework. Section~\ref{sec:exp_setup} and~\ref{sec:exp} present the experimental setup and the evaluations on the datasets we used. Finally, section ~\ref{sec:discussion} discusses an interesting finding we observed in our work.

\section{Related work} \label{sec:RW}

\minisection{Active learning.}  
Active Learning has been widely studied in various applications such as image classification \cite{snoek2015transfer,wu2018kdd,choi2021}, image retrieval \cite{barz2018retrieval}, image captioning \cite{Deng2018al}, object detection \cite{zolfaghari2019temporal}, and regression \cite{freytag2014influential,denzler2018bmvc}. 


Over the past two decades, several strategies have been proposed for sample query, which can be divided in three main categories: informativeness \cite{yang2015multi,gal2017icml,guo2010nips, gu2014modelchange,bengar2021deep}, representativeness \cite{falcao2015pr,sener2018active} and
hybrid approaches \cite{zhou2014hybrid,loog2018maxvariance}. A comprehensive survey of these frameworks and a detailed discussion can be found in \cite{settles2012active}.


Among all the aforementioned strategies, the informativeness-based approaches are the most successful ones, with uncertainty being the most used selection criteria used in both bayesian \cite{gal2017icml} and non-bayesian frameworks \cite{li2013adaptive}. In \cite{gal2017icml}, they obtain uncertainty estimates through multiple forward passes with Monte Carlo Dropout, but it is computationally inefficient for recent large-scale learning as it requires dense dropout layers that drastically slow down the convergence speed. More recently, \cite{ash2019deep} measures the uncertainty of the model by estimating the expected gradient length. On the other hand, \cite{Yoo_2019_CVPR,li2020learning} employ a loss module to learn the loss of a target model and select the images based on their output loss. 

Representativeness-based methods rely on selecting examples by increasing diversity in a given batch \cite{dutt2016active}. The Core-set technique  \cite{sener2018active} selects the samples by minimizing the Euclidian distance between the query data and labeled samples in the feature space. The Core-set technique is shown to be an effective representation learning method, however, its performance is limited by the number of classes in the dataset. Furthermore, Core-set, like other distance-based approaches, are less effective due to feature representation in high-dimensional spaces since p-norms suffer from the curse of dimensionality \cite{donoho2000high}. In a different direction, \cite{Sinha_2019_ICCV} uses an adversarial approach for diversity-based sample query, which samples the data points based on the discriminator's output, seen as a selection criteria. Following the same strategy, improved versions have been proposed in \cite{zhang2020sraal,kim2021tavaal}.

\minisection{Self-supervised learning.} 
In self-supervised learning, an auxiliary task is introduced. The data for this task should be readily available without the need for any human annotation. The auxiliary task allows to perform unsupervised learning and learn feature representations without the need of labels. Doersch et al.~\cite{doersch2015unsupervised} introduce the task of estimating the relative position of image regions. Other examples include coloring gray-scale images~\cite{zhang2016colorful}, inpainting~\cite{pathak2016context}, and ranking~\cite{liu2019exploiting}.

\begin{figure*}[h]
    \centering
    \includegraphics[width=.75\textwidth]{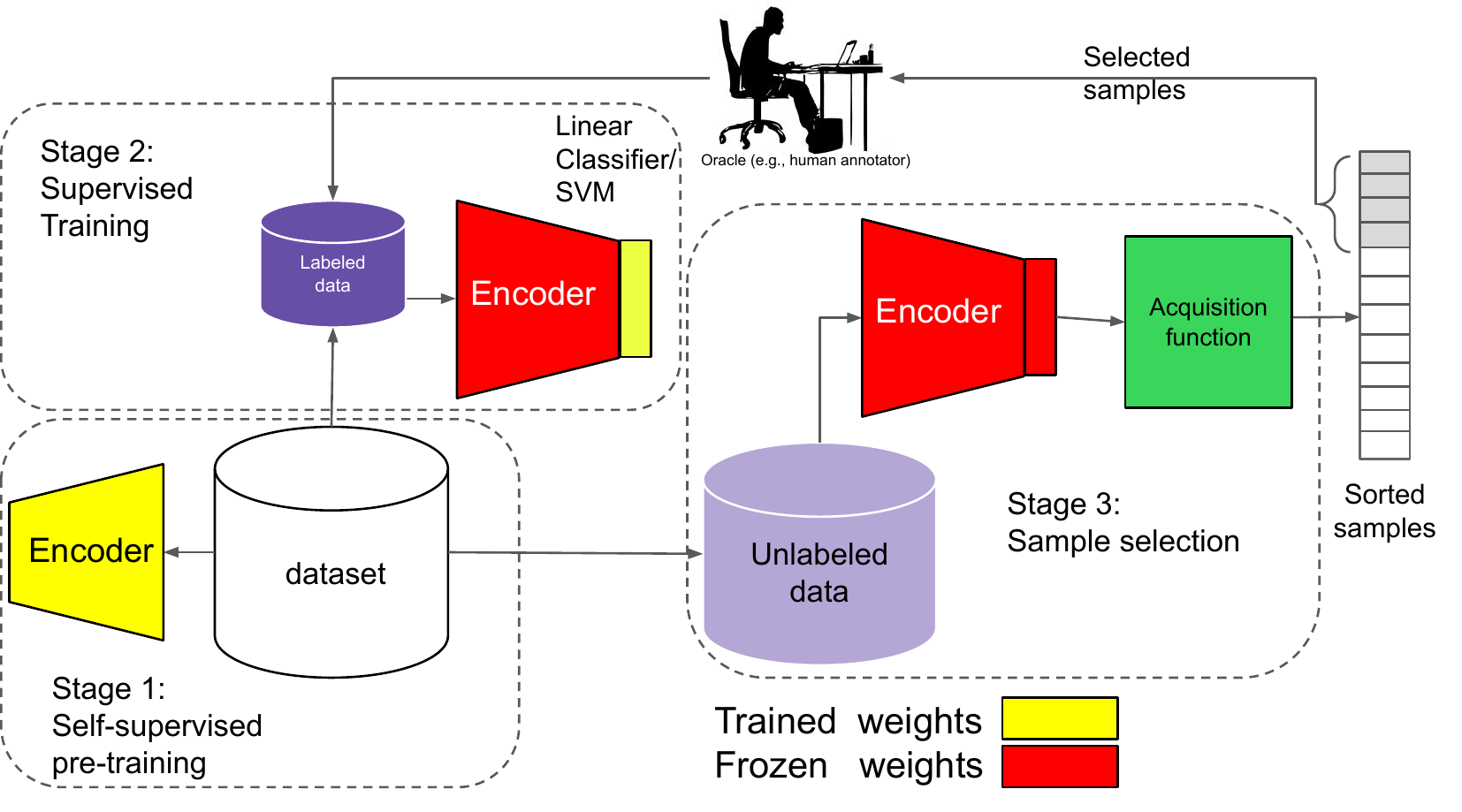}
    \caption{\small \textbf{Overview of active learning framework enhanced by self supervised pre-training.}
    The framework consists of 3 stages: (i) Self supervised model is trained on the entire dataset. (ii) Given the frozen backbone and few labeled data, a linear classifier or an SVM is fine-tuned on top of the features in supervised way. (iii) Running the model as inference on the unlabeled data and sort the samples from least to highest informative/representative via acquisition function. Finally the top samples are queried to oracle for labeling and added to labeled set. Stages (i) \& (ii) are repeated until the total labeling budget finishes.}
    \label{fig:AL_framework}
\end{figure*}

In recent years, self-supervised learning has seen a significant performance jump with the introduction of contrastive learning~\cite{chen2020SimCLR}, where representations are learned that are invariant with respect to several image distortions. Similar samples are created by augmenting an input image, while dissimilar are chosen by random. This connects to some extent unsupervised setting to previous contrastive methods used in metric learning~\cite{hadsell2006dimensionality,weinberger2006triplet}. To make contrastive training more efficient MoCo method~\cite{he2020MOCO} and the improved version~\cite{chen2020MOCOv2} use memory bank for learned embeddings what helps with an efficient sampling. This memory is kept in sync with the rest of the network during the training time by using a momentum encoder. Approach named SwAV~\cite{caron2020SwAV} use online clustering over the embedded samples. In this method negative exemplars are not defined. However, others cluster prototypes can play this role. Even more interesting are methods without any explicit contrastive pairs. BYOL~\cite{NEURIPS2020BYOL} propose asymmetric network by introducing of an additional MLP predictor between two branches' outputs. One of the branch is keep "offline" - updated by a momentum encoder. SimSiam~\cite{chen2021SimSiam} goes even further and presents a simplified solution without a momentum encoder. It comparably good to other methods and does not need a big mini-batch size. A follow up work of BarlowTwins~\cite{zbontar2021barlow} proposes as simple solution as SimSiam with the use of a different loss function - a correlation based one for each pair in current training batch. Here, negatives are implicitly assumed to be in each mini-batch. No asymmetry is used in the network at all, but a bigger embedding size and mini-batches are proffered in comparison to SimSiam.

Previous works that integrated Active Learning and Self-supervised learning include 
\cite{zhu2020contrastive,pathak2019self}.~\cite{zhu2020contrastive} proposes a query based graph AL method for datasets having structural relationships between the samples coming from few classes. In the context of exploration-driven agent, \cite{pathak2019self} uses Active Learning and Self-training to learn a policy that allows it to best navigate the environment space.

\section{Preliminaries} \label{sec:prelim}
The main objective of this paper is to evaluate and compare the effectiveness of active learning when combined with recent advances in self-supervised learning. For this purpose we have developed a framework that comprises two parts: self supervised pre-training and active learning (see Figure~\ref{fig:AL_framework}). Primarily, we train the self supervised model as the pretrained model on the unlabeled samples. Next, we use an initial labeled data to finetune a linear classifier on top of pre-trained model. Then we run active learning cycles using the fine-tuned model to select the most informative and/or representative samples and query them for labeling. Hence the original dataset becomes partially labeled. We ablate the self-supervised and active learning components to study their benefits.

We start pretraining our model with SimSiam \cite{chen2021SimSiam} self-supervised model. The model is based on siamese network trying to maximize the similarity between two augmentations of one image, subject to certain conditions for avoiding collapsing solutions. This enables us to obtain meaningful representations without using negative sample pairs. The rich representations could also potentially help the representative based active learning methods. 

In the remainder of this section we describe the two components of the experimental framework in detail.

\subsection{Active Learning} 
Given a large pool of unlabeled data $\mathcal{D_U}$ and a total annotation budget $B$, the goal is to select $b$ samples in each cycle to be annotated to maximize the performance of a classification model. In general, AL methods proceed sequentially by splitting the budget in several \emph{cycles}.
Here we consider the batch-mode variant~\cite{settles2012active}, which annotates $b$ samples per cycle, since this is the only feasible option for CNN training.
At the beginning of each cycle, the model is trained on the labeled set of samples $\mathcal{D_L}$.
After training, the model is used to select a new set of samples to be annotated at the end of the cycle via an \emph{acquisition function}. 
The selected samples are added to the labeled set $\mathcal{D}_L$ for the next cycle and the process is repeated until the annotation budget is spent. The acquisition function is the most crucial component and the main difference between AL methods in the literature. In the experiments we consider several acquisition functions including Informativeness \cite{dagan1995committee} and Representativeness based methods \cite{Sinha_2019_ICCV,sener2018active}.

\begin{figure}[t]
    \centering
    \includegraphics[width=\columnwidth]{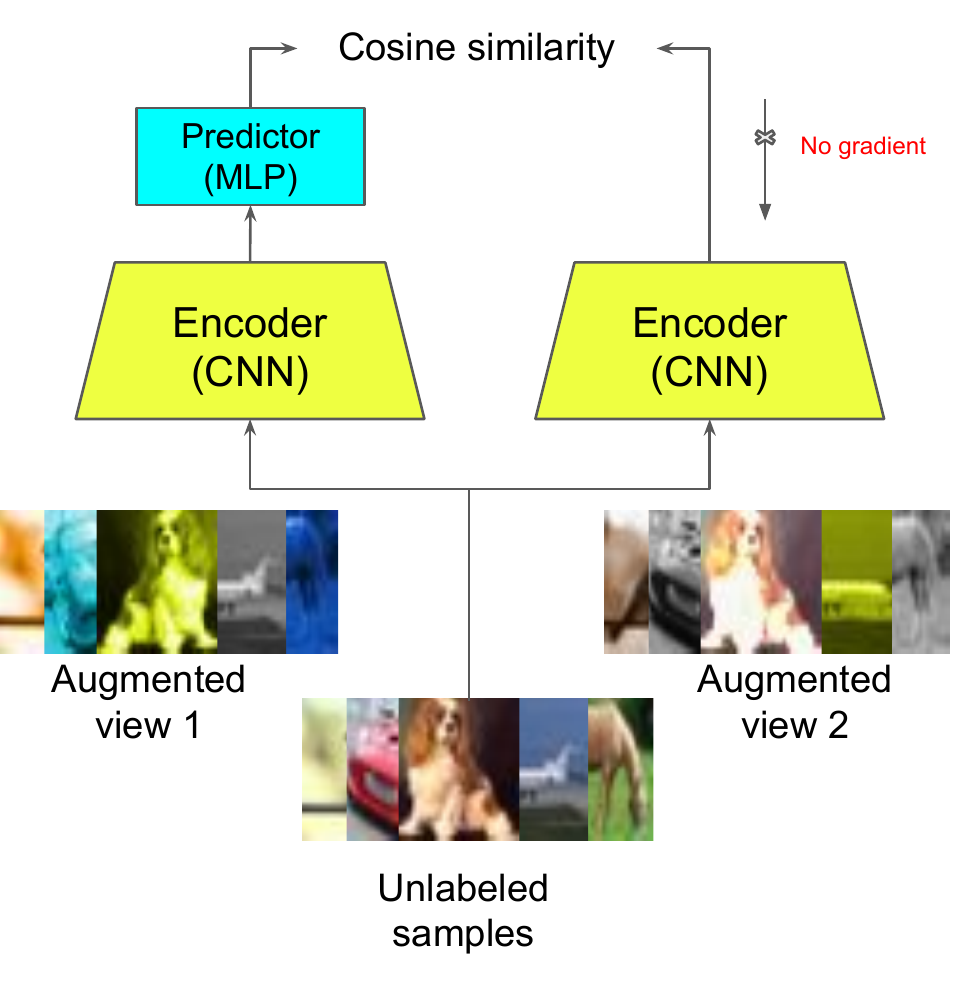}
    \caption{\small \textbf{SimSiam architecture} Two augmented views of one image are processed by the same encoder network (a backbone
    plus a projection MLP). Then a prediction MLP is applied on one
    side, and a stop-gradient operation is applied on the other side. The model maximizes the similarity between both sides.}
    \label{fig:simsiam}
\end{figure}

\subsection{Self-supervised Learning} 
In this section, we shortly introduce self-supervised learning without contrastive sampling and more particularly SimSiam~\cite{chen2021SimSiam}, the architecture we employ in this paper. 

For a given dataset $\mathcal{D}$, contrastive learning assumes sampling pairs of data points in order to create a good representation. Two main types of pairs are considered: \emph{semantically similar pairs} $(x, x^{+})$ -- provide an information about some form of close relation of data (based on labeled or unlabeled data); \emph{negative pairs} $(x, x^{-})$ -- in contrast to positives ones, two non-related samples are given. It is presumed that for a given $x$, $x^{-}$ is dissimilar to $x^{+}$. Then, contrastive losses~\cite{hadsell2006dimensionality, weinberger2006triplet} learn a new embedding space where a distance between positive pairs is smaller than negatives ones with some margin, e.g. $d(x, x^{+}) < d(x, x^{-}) + m$ for triplet loss~\cite{weinberger2006triplet}. That is a core of many metric learning methods~\cite{kaya2019dml_survey,musgrave2020realitycheck}, where existing labels are used for a semantic similarity check.

Contrastive learning is also often applied in self-supervised learning methods. These methods aim to learn a semantically rich feature representation without the need of any labels.
Different augmented views of the same image $x$ form positive samples, while augmentation of different ones provide negatives. This is the base of SimCLR~\cite{chen2020SimCLR} method. However, it's shown that methods without explicit negative sampling prove competitive performance as well, e.g. SimSiam~\cite{chen2021SimSiam} or BYOL~\cite{NEURIPS2020BYOL}. In such methods some additional architecture changes are usually applied, like using asymmetry with an additional predictor network as presented in Figure~\ref{fig:simsiam} for SimSiam. The main part is an encoder (CNN based network), learned end-to-end in an asymmetric Siamese architecture, where one branch got an additional predictor (MLP network) which outputs aims to be as close as possible to the other branch. The second branch is not updated in a backward propagation while training. For the similarity function a negative cosine distance is minimized given as: 
\newcommand{\dist}{\mathcal{D}}
\newcommand{\p}{{p}}  
\newcommand{\z}{{z}}  
\newcommand{\lnorm}[1]{\frac{#1}{\left\lVert{#1}\right\rVert _2}}
\newcommand{\lnormv}[1]{{#1}/{\left\lVert{#1}\right\rVert _2}}
\begin{equation}
    \mathcal{L} = \dist(\p_1, \z_2)/2 + \dist(\p_2, \z_1)/2
\label{eq:loss_simsiam}
\end{equation}
\begin{equation}
\dist(\p_1, \z_2) = - \lnorm{\p_1}{\cdot}\lnorm{\z_2},
\label{eq:dist_cosine}
\end{equation}
where $z_1$, $z_2$ are encoded values respectively for $x1$ and $x2$ -- two different augmented views of the same image $x$. $p_1$ and $p_2$ are encoded values additionally passed by a predictor network. There is no contrastive term in this approach, only the similarity is checked and enforced during learning. In SimSiam, besides simplicity, neither negatives mining nor large mini-batches are needed which significantly reduces the GPU requirements. This makes it a good fit for the evaluation proposed in this paper.

\section{Experimental Setup} \label{sec:exp_setup}  
To study the influence of the initial model, various amounts of initial labeled data and budget sizes are evaluated. For the initial labeled set, we considered $1\%$, $2\%$ and $10\%$ of the entire dataset that are uniformly selected from all classes at random. For one of the datasets we also evaluate $0.1\%$ and $0.2\%$ budget sizes. Before starting the active learning cycles we train the self-supervised model. Then we use the backbone as encoder from SimSiam architecture, freeze the weights and train a linear classifier or SVM on top of the backbone so we only finetune the last layer. At each cycle we start training either from scratch or, in case of self-training, we start from the pretrained self-supervised backbone. We train the model in $c$ cycles until the total budget is exhausted. In each experiment the budget per cycle is equal to initial labeled set.

\minisection{Datasets.}
To evaluate various methods, we use CIFAR10 and CIFAR100 \cite{krizhevsky2012learning} datasets with 50K images for training and 10K for testing. CIFAR10 and CIFAR100 have 10 and 100 object categories respectively and an image size of 32$\times$32.
To evaluate the scalability of the methods we evaluate on Tiny ImageNet dataset \cite{le2015tiny} with 90K images for training and 10K for testing. There are 200 object categories in Tiny ImageNet with an image size of 64$\times$64. 

\minisection{Data Augmentation}
We use different augmentation policies for self supervised pre-training and supervised finetuning. \cite{zoph2020rethinking} discusses how self-training outperforms normal pre-training in terms of stronger augmentation. 
For the self-training similar to \cite{chen2021SimSiam} we used Geometric augmentations \cite{wu2018unsupervised}: RandomResizedCrop with scale in [0:2; 1:0] and RandomHorizontalFlip. Color augmentation is ColorJitter with \{brightness, contrast, saturation,
hue\} strength of \{0.4, 0.4, 0.4, 0.1\} with an applying
probability of 0.8, and RandomGrayscale with an applying
probability of 0.2. Blurring augmentation \cite{chen2020SimCLR} has a Gaussian kernel with std in [0:1; 2:0]. For the supervised training we used the conventional RandomResizedCrop with scale [0.08, 1.0] and RandomHorizontalFlip.
\paragraph{Baselines.}
For the evaluation baselines we compared with Random sampling and several informative and representative-based approaches including Entropy sampling, KCenterGreedy, VAAL and SVM Min Margin. Below we describe the details of the methods we used.

\emph{Entropy}~\cite{dagan1995committee} is an information theory measure that captures the average amount of information contained in the predictive distribution, attaining its maximum value when all classes are equally probable. Entropy sampling selects the most uncertain samples with highest entropy.

As a prominent representative method we evaluate \emph{KCenterGreedy}, which is a greedy approximation of KCenter problem  also  known  as  min-max  facility location problem \cite{wolf2011facility}. The method selects samples having maximum distance from the nearest labeled samples in the embedding space. We compute the embeddings by running the self-trained model on unlabeled samples.

\emph{VAAL} \cite{Sinha_2019_ICCV} is one of state-of-art methods that uses a variational autoencoder to map the distribution of labeled and unlabeled data to a latent space. A binary adversarial classifier  is trained to predict if an image belongs to the labeled or the unlabeled pool. The unlabeled images which the discriminator classifies with lowest certainty as belonging to the labeled pool are considered to be the most representative. We used their official code and adapted them into our code to ensure an identical setting.
To adapt VAAL for the self-training experiment we initialized and froze the backbone of the task learner.

\emph{SVM Min Margin} \cite{tong2001support} learns a linear SVM on the existing labeled data and chooses the samples that are closest to the decision boundary. To generalize SVM for the multi-class classification problem we adopt it by querying the samples that reside in margin  area of decision boundaries.

\paragraph{Implementation details.}
Our method is implemented in PyTorch \cite{paszke2017automatic}. We train Resnet18 \cite{He_2016} that is widely used on CIFAR10 and CIFAR100 datasets.    
For the self-supervised training, the models are trained with SGD optimizer with momentum 0.9 and base learning rate of 0.03. As in \cite{chen2021SimSiam} we train models for 800 epochs with batch-size of 512. We use a weight decay of 0.0001 for all parameter layers, including the BN scales and biases, in the SGD optimizer.

Given the pre-trained network, we train a supervised linear classifier on frozen features, which are from ResNet’s global average pooling layer. The linear classifier training uses base lr=30 with a cosine decay schedule for 100 epochs, weight decay = 0, momentum=0.9, batch size=256 with SGD optimizer.

To implement the SVM for the Min Margin method we used scikit learn python package \cite{scikit-learn} with linear kernel and set the regularization parameter to 5 in the experiments. To handle the multi-class problem, a one-vs-the-rest classification scheme is chosen.

\section{Experiments} \label{sec:exp} 
To evaluate active learning methods we consider several scenarios in the initial labeled set and budget sizes. For the simplicity we refer to lower than $2\%$ budget sizes as low budget regimes. In this section we inspect the contribution of self-supervised pre-training in active learning.

\paragraph{Performance on CIFAR10.}
\begin{figure}[t]
    \centering
    \includegraphics[width=\columnwidth]{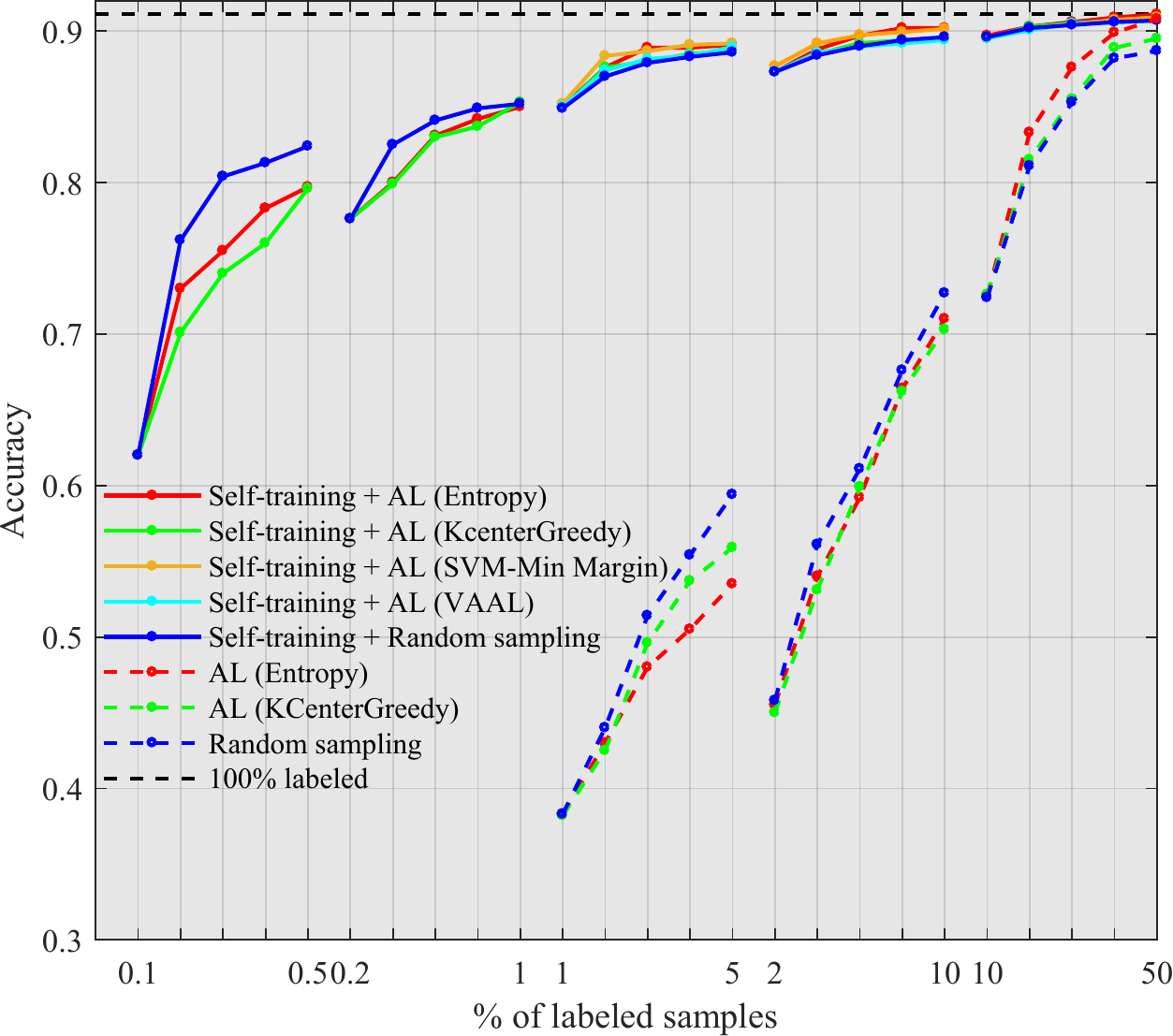}
    \caption{\small \textbf{AL performance on cifar10} performance comparison between the addition of self-training to AL methods (solid lines) and AL methods (dashed lines). The initial and per cycle budget are equal in all the curves.}
    \label{cifar10}
\end{figure}

Figure \ref{cifar10} shows active learning results on CIFAR10 dataset. The initial and per cycle budgets are $0.1\%$, $0.2\%$,$1\%$, $2\%$ and $10\%$ of labeled data. The evaluated methods are divided into two groups: (i) methods using self-supervised pre-training represented by solid lines. (ii) Methods using models trained from scratch represented by dashed lines. As can be seen, self-training substantially improves all the sampling methods. In particular at the low budget regime, self-training drastically reduces the required labeling. 
Both types of methods achieve almost the full performance after labeling $50\%$ of data that closes the gap between the self-supervised and supervised methods. The exact numbers are in Table \ref{tab:table_overall}.
From the active learning perspective, Random sampling outperforms AL methods when the budget is less than $1\%$. However from $1\%$ budget onward, AL + self-training methods transition to higher performance compared to Random sampling with self-training. For AL methods, trained from scratch, this transition happens after labeling $10\%$ of data.
Among AL methods with self-training, Entropy as informativeness method outperforms KCenterGreedy and VAAL. Note that the greatest active learning gain as a result of using self-training occurs after labeling $30\%$ providing $20\%$ less annotation that is equivalent to 10000 less labeling. 

\paragraph{Performance on CIFAR100.}
\begin{figure}[t]
    \centering
    \includegraphics[width=\columnwidth]{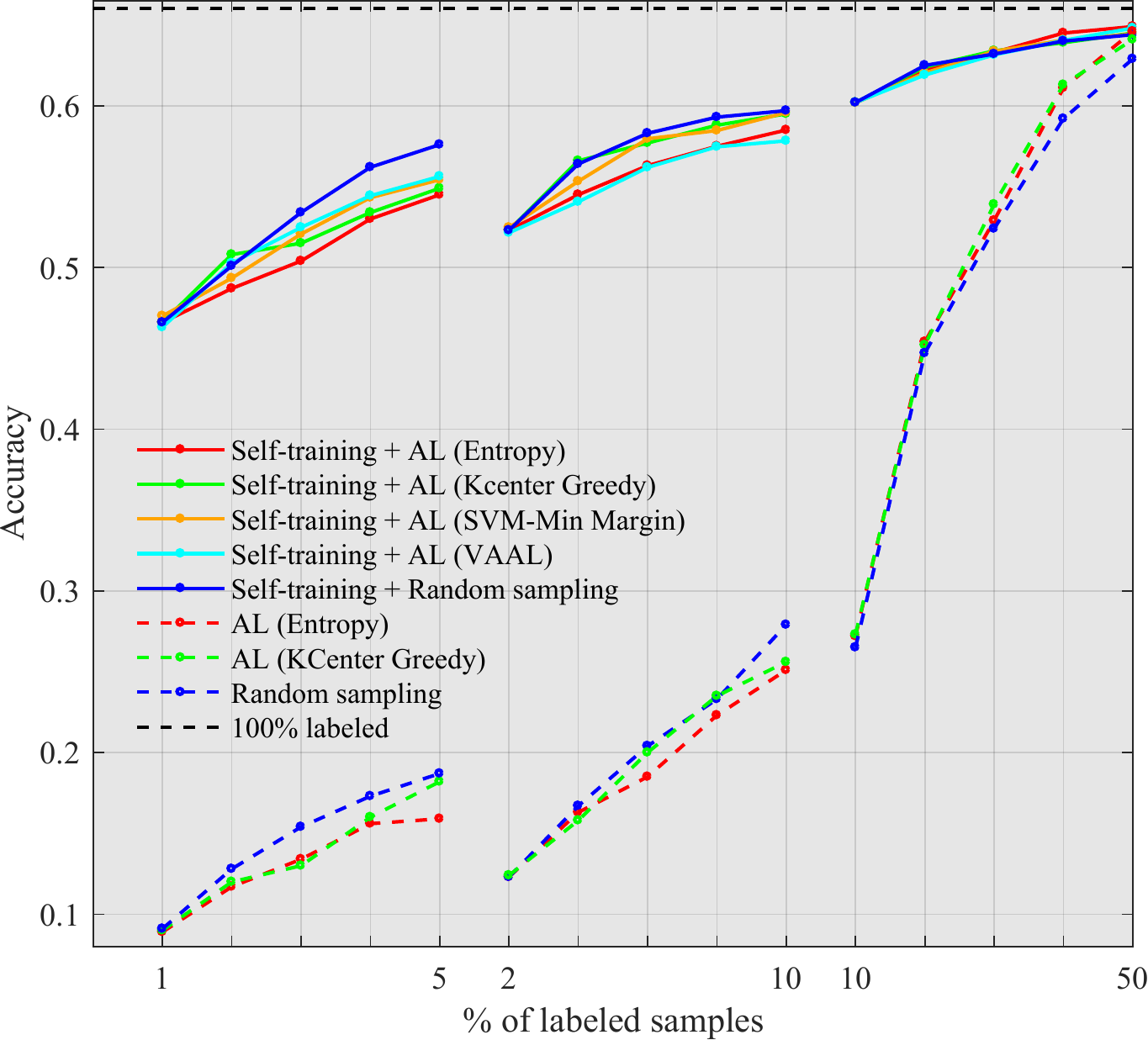}
    \caption{\small \textbf{AL performance on cifar100} performance comparison between the addition of self-training to AL methods (solid lines) and AL methods (dashed lines). The initial and per cycle budget are equal in all the curves.}
    \label{cifar100}
\end{figure}
Figure \ref{cifar100} presents active learning results on CIFAR100 dataset. The three set of curves correspond to three initial and per cycle budget sizes: $1\%$, $2\%$ and $10\%$. Solid lines represent AL methods using self-supervised training. While dashed curves correspond to algorithms trained from scratch. As can be seen, self-training dramatically improves the methods without self training. In the low budget regime, self-training significantly reduces the required labeling. While AL methods w/o self-training achieve comparable performance to self-trained counterparts as we approach to $50\%$ labeled data, meaning that the impact of self supervised pre-training diminishes when the budget increases. See Table \ref{tab:table_overall} for detailed numbers. This can also be due to reaching almost the full performance. On CIFAR100, Random sampling outperforms Active learning methods under low budget regardless of using self-training. None of the studied methods foresee a regime where the labeling budget is small, for example, labeling lower than $10\%$. Among the AL methods with self-training, representative-based methods perform better than Entropy as informative-based in low budget. On CIFAR100, the active learning gain of using self-training appeared almost after labeling $40\%$ of dataset resulting in $10\%$ less annotation that is equivalent to 5000 less labeling. 

\paragraph{Performance on Tiny ImageNet.}
Tiny ImageNet is a challenging dataset in terms of diversity of classes. Active learning results on this dataset is presented in Figure \ref{cifar100}. Similar to CIFAR100, the three set of curves correspond to $1\%$, $2\%$ and $10\%$ budget per cycle. Solid lines represent AL methods with self-supervised pre-training and dashed lines correspond to methods trained from scratch. As in other datasets, Self-training drastically reduces the required labeling in low budget scheme. As the labeling increases to $50\%$ AL methods approach the performance of self-trained counterparts. However, unlike CIFAR datasets, AL methods require more than $50\%$ labeling to close the performance gap they have from self-trained counterparts.
Among the methods using self training, Random sampling shows superior performance. However, increasing labeled data reduces performance gap from the AL methods. 
For AL methods w/o self-training, the labeling budget is required to exceed $10\%$ to improve upon Random sampling. 
In general, active learning fails to perform well under low budget regardless of using self-training. Again AL methods are not designed for low budget regime. Unless the model is trained from scratch with greater than $10\%$ labeling budget, we observe no improvement with the usage of Active learning. 

\begin{figure}[t]
    \centering
    \includegraphics[width=\columnwidth]{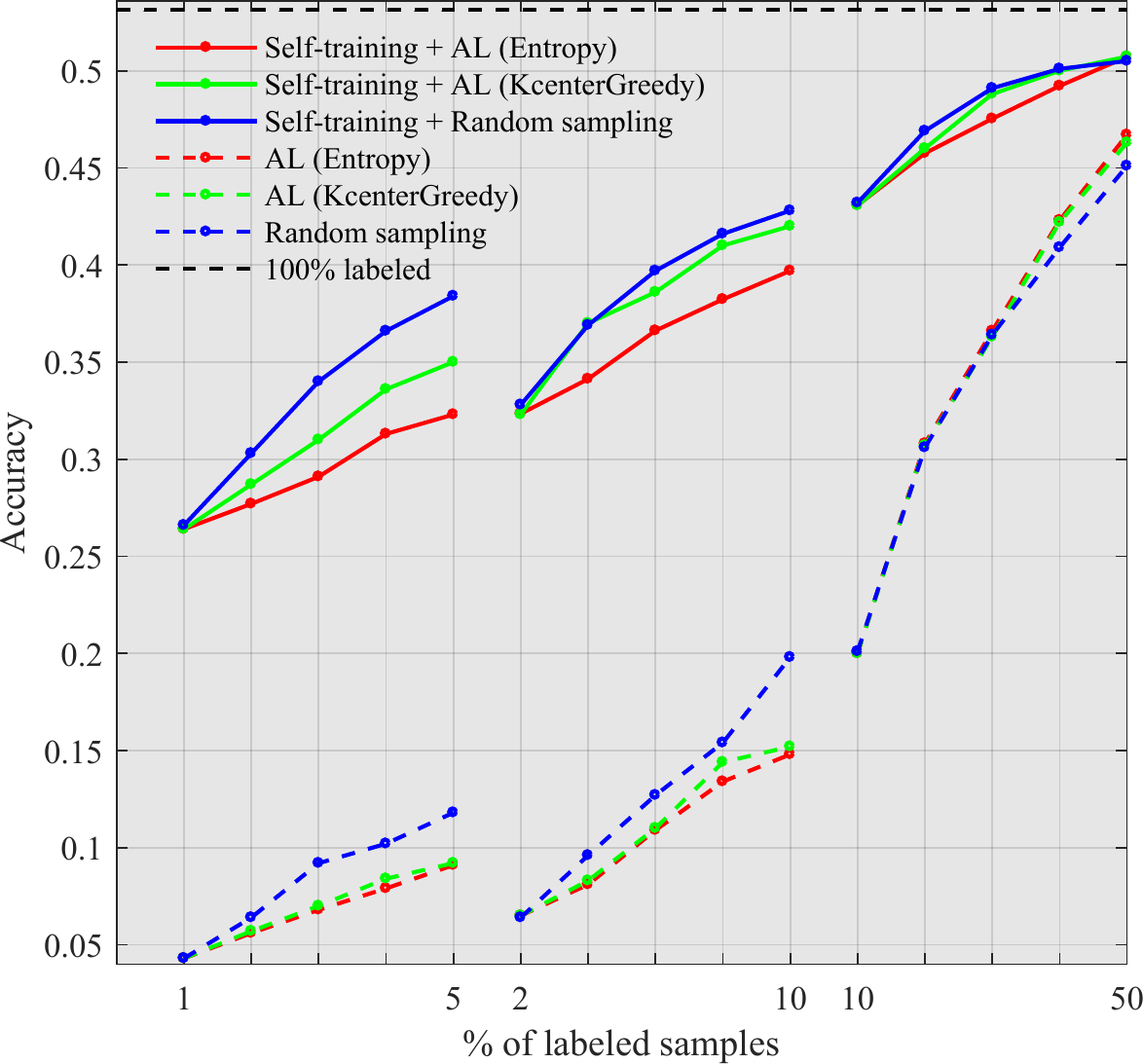}
    \caption{\small \textbf{AL performance on Tiny ImageNet} performance comparison between the addition of self-training to AL methods (solid lines) and AL methods (dashed lines). The initial and per cycle budget are equal in all the curves.}
    \label{tinyImageNet}
\end{figure}
\section{Discussion}\label{sec:discussion}
\begin{figure}[t]
    \centering
    \includegraphics[width=\columnwidth]{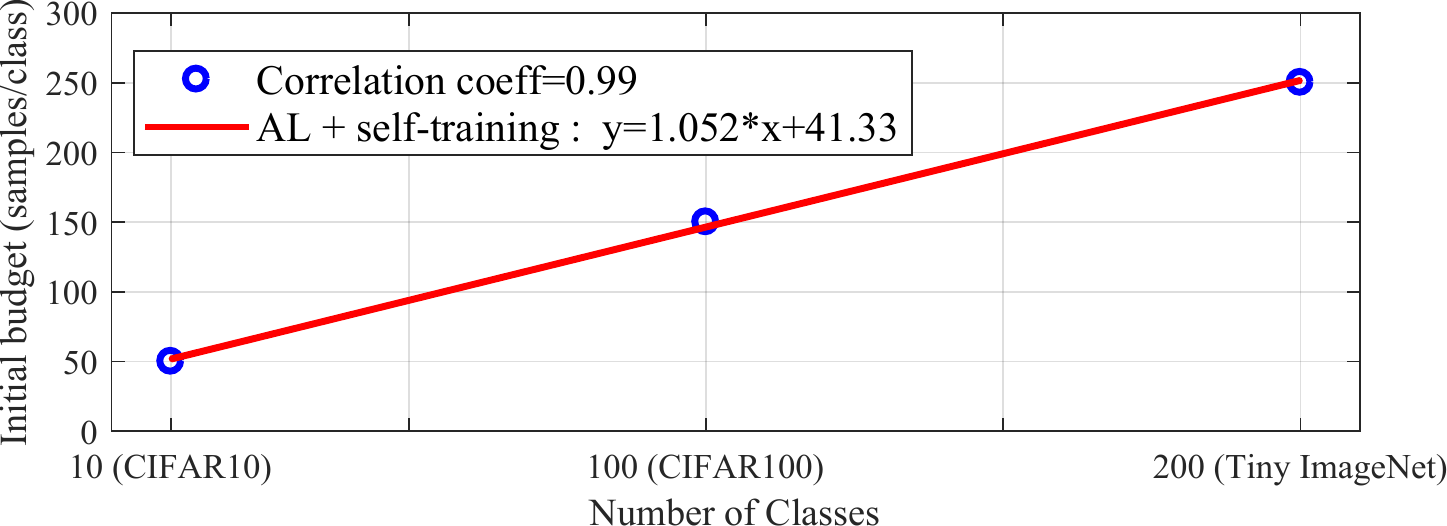}
    \caption{\small \textbf{Correlation between number of samples per class required for AL and number of classes in the datasets.} Above these budgets, AL outperforms Random sampling in the self-supervised setting.}
    \label{correlation}
\end{figure}

The experiments in the previous section demonstrated that active learning methods enhanced by self-training do not work well in all budget schemes. However, it might be possible to estimate budgets above which the AL methods outperform Random sampling. Our experiments on three object recognition datasets show that there's a strong correlation (corr. coeff=0.99) between the number of samples per class required for AL and the number of classes in a datasets. Figure \ref{correlation} presents the thresholds for the budget required for active learning to improve upon Random sampling when uses self-training. This is one interesting finding we observed which can provide a guideline based on the number of classes in a dataset to decide with a certain labeling budget whether it's beneficial to use active learning.  

\begin{table}[t]
    \centering
    \resizebox{\columnwidth}{!}{
    \begin{tabular}{ll ccccc}
        \toprule
         \multirow{2}{*}{} & \multirow{2}{*}{\bf Methods}&
         \multicolumn{2}{c}{\bf Datasets} \\
         & & CIFAR10 & CIFAR100 \\
        \midrule
        \multirow{2}{*}{\bf AL w/o Self-training} & Entropy & 0.908 & 0.646 \\
           & KCenterGreedy &  0.895 & 0.641 \\
        \midrule
        \multirow{4}{*}{\bf AL + Self-training } & Entropy & 0.911 & 0.649 \\
             & SVM Min Margin &  0.909& 0.644 \\
             & VAAL  & 0.907 & 0.648 \\
             & KCenterGreedy & 0.909 & 0.645 \\
         \bottomrule
    \end{tabular}}
    \vspace{2mm}
    \caption{\small  \textbf{Performance of AL methods with and without Self-training at $50\%$ labeling.} For the high labeling budget, the gap between the performances of AL and AL+ Self-training is diminished.}
    \label{tab:table_overall}
    \vspace{-4mm}
\end{table}

\section{Conclusions}
This paper analyzed active learning and self supervised approaches independently and unified to investigate how they can benefit from each other. Our experiments demonstrated that self-training is way more efficient than active learning at reducing the labeling effort. Besides, for a low labeling budget, active learning brings no benefit to self-training. Finally, the combination of active learning and self-training is beneficial only when the labeling budget is high. The performance gap between active learning with and without self-training diminishes as we approach to the point where almost half of the dataset is labeled.

\textbf{Acknowledgement} We acknowledge the support of the project  PID2019-104174GB-I00 (MINECO, Spain) and the CERCA Programme of Generalitat de Catalunya.

{\small
\bibliographystyle{ieee_fullname}
\bibliography{egbib}
}

\end{document}